\documentclass[conf]{new-aiaa}
\usepackage[utf8]{inputenc}

\usepackage{graphicx}
\usepackage{amsmath}
\usepackage{subcaption}
\usepackage{wrapfig}
\usepackage{bbm}
\usepackage{soul}

\usepackage{algorithm}
\usepackage{algorithmic}
\usepackage{multirow}
\usepackage[dvipsnames]{xcolor}
\usepackage{tikz}

\usepackage[version=4]{mhchem}
\usepackage{siunitx}
\usepackage{longtable,tabularx}
\newcommand\copyrighttext{%
  \footnotesize Copyright~\textcopyright~2024 by the American Institute of Aeronautics and Astronautics, Inc. All rights reserved.
}
\newcommand\copyrightnotice{%
\begin{tikzpicture}[remember picture,overlay]
\node[anchor=south,yshift=10pt,align=center] at (current page.south) {\fbox{\copyrighttext}};
\end{tikzpicture}%
}

\title{Verification and Validation of a Vision-Based Landing System for Autonomous VTOL Air Taxis}

\author{
Ayoosh Bansal\footnote{Authors contributed equally.}\footnote{Postdoctoral Research Associate, Computer Science, 201 North Goodwin Avenue, Urbana, IL 61801}}\affil{University of Illinois at Urbana-Champaign, Urbana, Illinois, 61801}

\author{
Duo Wang*\footnote{Postdoctoral Research Associate, Mechanical Engineering Department, 1664 N. Virginia Street, Reno, NV 89557}
\footnote{Visiting Scholar at UIUC, Mechanical Science \& Engineering Department, 105 S Mathews Ave, Urbana, IL 61801}}
\affil{University of Nevada, Reno, Nevada, 89557}

\author{
Mikael Yeghiazaryan*\footnote{Visiting Scholar at UIUC, Mechanical Science \& Engineering Department, 105 S Mathews Ave, Urbana, IL 61801}, Yangge Li\footnote{Graduate Student, Electrical and Computer Engineering, 306 N. Wright St., Urbana, IL 61801}, Chuyuan Tao\footnote{Graduate Student, Mechanical Science \& Engineering Department, 105 S Mathews Ave, Urbana, IL 61801}}
\affil{University of Illinois at Urbana-Champaign, Urbana, Illinois, 61801}

\author{Hyung-Jin Yoon\footnote{Assistant Professor, Mechanical Engineering Department, 1 William L Jones Dr, Cookeville, TN 38505}}
\affil{Tennessee Technological University, Cookeville, Tennessee, 38505}

\author{Prateek Arora\footnote{Graduate Student, Department of Computer Science \& Engineering, 1664 N. Virginia Street, Reno, NV 89557}, Christos Papachristos\footnote{Assistant Professor, Department of Computer Science \& Engineering, 1664 N. Virginia Street, Reno, NV 89557}, Petros Voulgaris\footnote{Professor, Mechanical Engineering Department, 1664 N. Virginia Street, Reno, NV 89557}}
\affil{University of Nevada, Reno, Nevada, 89557}

\author{
Sayan Mitra\footnote{Professor, Electrical and Computer Engineering, 306 N. Wright St., Urbana, IL 61801},
Lui Sha\footnote{Donald B. Gillies Chair in Computer Science, Computer Science, 201 North Goodwin Avenue, Urbana, IL 61801, USA.}}
\author{Naira Hovakimyan\footnote{Professor, Mechanical Science \& Engineering Department, 105 S Mathews Ave, Urbana, IL 61801}}
\affil{University of Illinois at Urbana-Champaign, Urbana, Illinois, 61801}

\begin{document}

\maketitle

\copyrightnotice

\begin{abstract}

Autonomous air taxis are poised to revolutionize urban mass transportation.
A key challenge inhibiting their adoption is ensuring the safety and reliability of the autonomy solutions that will control these vehicles.
Validating these solutions on full-scale air taxis in the real world presents complexities, risks, and costs that further convolute the challenge of ensuring safety and reliability of these autonomous vehicles.
Verification and Validation (V\&V) frameworks play a crucial role in the design and development of highly reliable systems by formally verifying safety properties and validating algorithm behavior across diverse operational scenarios.
Advancements in high-fidelity simulators have significantly enhanced their capability to emulate real-world conditions, encouraging their use for validating autonomous air taxi solutions, especially during early development stages.
This evolution underscores the growing importance of simulation environments, not only as complementary tools to real-world testing but as essential platforms for evaluating algorithms in a controlled, reproducible, and scalable manner.

This work presents a V\&V framework for a vision-based landing system for air taxis with vertical take-off and landing (VTOL) capabilities.
Specifically, we use Verse, a tool for formal verification, to model and verify the safety of the system by obtaining and analyzing the reachable sets.
To conduct this analysis, we utilize a photorealistic simulation environment.
The simulation environment, built on Unreal Engine, provides realistic terrain, weather, and sensor characteristics to emulate real-world conditions with high fidelity.
To validate the safety analysis results, we conduct extensive scenario-based testing to assess the reachability set and robustness of the landing algorithm in various conditions.
This approach showcases the representativeness of high-fidelity simulators, offering an effective means to analyze and refine algorithms before real-world deployment.

\end{abstract}

\section{Introduction}

The rapid development of autonomous aerial vehicles has significantly advanced modern aviation, enabling advancements in applications such as surveillance, delivery, agriculture, and disaster response. Vertical Take-Off and Landing (VTOL) aircrafts are a class of aerial vehicles capable of ascending and descending vertically, eliminating the need for runways. This capability is achieved through various designs, including rotorcraft, tiltrotor aircraft, and various emerging electric VTOL models. These systems promise efficiency and mobility, but ensuring their reliability remains a critical challenge, particularly for safety-critical operations such as landing in a cluttered urban environment. Autonomy algorithms, which govern these systems, are required to function reliably across a wide range of operational conditions, making rigorous Verification and Validation (V\&V) frameworks indispensable. These frameworks systematically verify safety properties and validate the performance of these algorithms, ensuring they can be deployed with high confidence in real-world environments.

Robustness of the autonomy algorithms is integral to the safe and efficient operation of autonomous VTOL aircrafts~\cite{MIT_AutonomousVTOL, wei2024autonomous, feary2023evaluation}.
Autonomy systems typically comprise of three major components: perception, path planning, and control. Perception algorithms enable the vehicle to interpret its surroundings by processing data from onboard sensors such as cameras, LiDAR, and radar to identify suitable landing zones and assess environmental features~\cite{tepylo2023public}. Path planning algorithms compute feasible and optimized trajectories that consider vehicle dynamics, environmental constraints, and mission objectives~\cite{chen2022autonomous}. Control algorithms ensure the precise execution of these trajectories, maintaining stability and robustness throughout complex maneuvers such as vertical landing~\cite{ducard2021review, bauersfeld2021mpc}. The seamless integration of these components is essential for achieving reliable performance, as failures in any one of them can compromise the overall safety of the system. This interconnected nature of autonomy algorithms underscores the need for holistic development and testing to ensure their reliability across diverse and dynamic operational scenarios.

Hybrid system verification offers a mathematical framework to capture and analyze the complex interactions of continuous dynamics and discrete transitions that characterize autonomous systems. The increasing complexity of such systems, driven by advanced sensors, perception modules, and controllers, has spurred the development of tools to analyze both linear and nonlinear system behaviors. Validation, on the other hand, addresses the practical aspects of autonomy algorithm performance by testing them in simulated or real-world environments. It aims to assess whether these systems meet operational requirements when subject to realistic conditions, including sensor noise, environmental variability, and actuator imperfections~\cite{sargent2010verification}. Together, verification and validation provide a complementary approach to assessing autonomy algorithms, ensuring both theoretical soundness and practical reliability.

Current V\&V practices for autonomy algorithms rely on simulation environments due to their cost-effectiveness and controllability. Simulators enable repeatable experiments, controlled scenario design, and accelerated testing.
However, inevitably, simulation environments cannot capture the full complexity and details of the real-world, leading to the simulation-to-reality (sim-to-real) gap.
Sim-to-real gap here refers to the differences in an autonomy's solution performance within the simulated environment versus the real-world, specifically, the autonomy algorithms that perform well in simulations may fail to achieve the same level of reliability in real-world applications~\cite{jakobi1995noise,salvato2021characterization}.
This gap arises from factors such as incomplete sensor modeling, unrealistic environmental interactions, and oversimplified dynamics in the simulation environment. The persistence of this gap poses challenges for safety-critical applications, e.g., autonomous landing, where small inaccuracies can lead to significant consequences.

Recent advances in simulation technologies, particularly those leveraging 3D rendering engines like Unreal Engine~\cite{EpicUE}, have mitigated some of the limitations of traditional simulation environments by providing high-fidelity virtual settings. CARLA~\cite{dosovitskiy2017carla}, a simulation platform originally designed for ground vehicles and built on Unreal Engine, has been extended to support aerial vehicles~\cite{bansal2024synergistic}. This extension enables the testing of perception and control algorithms for air vehicles in realistic simulated environments. However, despite these advancements, significant challenges persist in achieving the level of fidelity required to close the sim-to-real gap. Addressing these challenges necessitates systematic efforts to enhance simulation realism, ensure algorithm robustness across diverse scenarios, and streamline the transfer of autonomy solutions from simulation to real-world applications.

In this work, we focus on the V\&V of vision-based landing algorithms for VTOL aircraft within a photorealistic simulation environment. Vision-based algorithms present unique validation challenges due to their reliance on high-quality sensor data and sensitivity to unpredictable factors such as lighting, texture, and occlusions. Our approach utilizes the \textit{Verse}~\cite{li2023verse} library to formally verify safety properties via reachability analysis, ensuring that the landing algorithm adheres to safety constraints under diverse conditions. The simulation environment offers a detailed representation of terrain, sensor characteristics, and environmental variations, facilitating realistic validation of perception and planning components. Through multiple experimental settings, we assess the landing algorithm's robustness and analyze the vehicle's reachable set, providing a rigorous evaluation of its performance.
By focusing on a photorealistic simulation environment, we provide insights into the strengths and limitations of current V\&V methods and lay the groundwork for future integration of these algorithms into hardware platforms.
This integration will improve the overall V\&V process, ensuring the reliability and safety of autonomous aerial systems in the real-world.

The key contributions of this paper are:
\begin{enumerate}
    \item We demonstrate the application of formal verification techniques using \textit{Verse}, providing a detailed analysis of safety properties of our landing algorithms and reachability under varying conditions.
    \item We present a validation framework that incorporates extensive scenario testing. Our work highlights its potential for bridging the sim-to-real gap and guiding the deployment of future hardware systems.
\end{enumerate}

By addressing these challenges, this study contributes to advancing the reliability and safety of autonomy algorithms, facilitating their deployment in aviation systems with increased confidence.

\section{Literature Review}

Autonomous operations of VTOL Unmanned Aerial Vehicles (UAVs) rely heavily on robust V\&V processes, advanced autonomy algorithms, and high-fidelity simulation environments. These elements are critical for ensuring reliable and safe performance enabling the future deployment of autonomy algorithms in real-world applications. In this context, V\&V efforts are specifically aimed at verifying the algorithms underlying autonomous decision-making, ensuring their robustness and correctness in various operational scenarios.
This section reviews related works and their limitations, highlighting the gaps that motivate the contributions made in this work.

\textbf{Tiltrotor VTOL UAV:} UAVs are generally classified into multirotor, fixed-wing, and hybrid VTOL types, each offering distinct operational advantages and limitations. \textit{Multirotor UAVs}, such as quadcopters, are capable of vertical take-off, landing, and hovering, making them suitable for tasks like aerial photography and surveillance in confined areas. However, they typically have limited flight endurance and speed due to the high energy consumption required to maintain lift~\cite{qin2023research}. \textit{Fixed-wing UAVs} resemble traditional airplanes, utilizing wings to generate lift, which allows for longer flight endurance and higher speeds, making them ideal for missions like large-area surveillance. Nonetheless, they require runways or launch systems for take-off and landing and lack the ability to hover~\cite{keane2017small}.

\textit{Hybrid VTOL UAVs} combine the vertical take-off and landing capabilities of multirotors with the efficient forward flight of fixed-wing aircraft, offering operational flexibility in diverse environments. This hybrid capability enables hybrid VTOL UAVs to perform missions that demand both hovering and long-range flight without the need for runways. The configuration of hybrid VTOL aircraft can vary, for example, tiltrotor (e.g., Bell~V-280~\cite{lopez2021bell}) and tilt-wing (e.g., NASA~GL-10~\cite{mcswain2017greased}).
A notable example of a hybrid VTOL UAV is the MiniHawk-VTOL~\cite{carlson2021minihawk, minihawk_github,robowork_aerial_robotics}, whose small scale is particularly suitable for testing autonomy algorithms. This rapidly prototyped tricopter/fixed-wing hybrid aircraft is designed for autonomous operations, featuring solar-recharge capability to support extended missions without the need for physical intervention. The MiniHawk-VTOL exemplifies the practical application of hybrid VTOL technology, combining efficient flight performance with operational flexibility. In this work, we utilize the MiniHawk-VTOL platform to explore and validate autonomy algorithms, leveraging its design to address challenges inherent in VTOL UAV operations.

\textbf{V\&V for autonomy algorithms:} Much of the existing research in autonomy algorithms (perception, path planning and control) has been developed primarily for multirotor and fixed-wing platforms, with limited direct application to hybrid VTOL configurations. \textit{Perception algorithms}, which are minimally influenced by the characteristics of hybrid VTOL UAVs, can leverage established methods from multirotor and fixed-wing UAVs. For instance, vision-based systems for obstacle avoidance and target tracking, commonly used in multirotors~\cite{cheng2017autonomous, wu2021vision}, and sensor fusion techniques for situational awareness in fixed-wing UAVs~\cite{kong2015ground,tian2021wind}, can be adapted to hybrid VTOL platforms without significant modifications. \textit{Path planning} and \textit{control} algorithms, however, require greater customization due to the unique dynamics of hybrid VTOL UAVs, which must transition seamlessly between hovering and forward flight. Algorithms like A* and its variants~\cite{cai2019path, zhou2022uav}, developed for multirotor and fixed-wing platforms, provide a foundation for hybrid VTOL path planning but often require further refinement to account for operational scenarios. Similarly, control strategies such as PID controllers and model predictive control (MPC) are insufficient on their own for hybrid VTOLs, which demand hybrid approaches capable of managing complex transitions between flight modes~\cite{bauersfeld2021mpc}. {Some state-of-the-art controllers require direct control of motor thrust and torque to achieve precise trajectory tracking and adaptability under complex conditions. However, in outdoor hardware experiments, directly controlling motor thrust and torque is typically impractical and unsafe due to the high risk of crashing the vehicle. To mitigate these risks, this work adopts a 3D A* planner for path planning and utilizes mature and conservative low-level PID controllers in ArduPilot flight stack~\cite{ardupilot}. This setup limits the control input to waypoints, providing a robust and reliable approach for future outdoor experiments while reducing the likelihood of hardware failures.}

Traditional V\&V techniques, often designed for deterministic and linear systems, struggle with the stochastic nature and adaptive behaviors of autonomy algorithms, particularly in hybrid VTOL UAVs where transitions between hover and forward flight modes introduce non-linear dynamics and unpredictable environmental interactions~\cite{redfield2017verification}. Tools like \textit{Hylaa}~\cite{bak2017hylaa}, \textit{DryVR}~\cite{fan2017dryvr}, and \textit{CORA}~\cite{Althoff2015a} have enabled the verification of systems with thousands of continuous variables, addressing challenges such as scalability and computational efficiency. Building on these advancements, the \textit{Verse} library~\cite{li2023verse} extends the capabilities of hybrid system verification by enabling modeling and safety analysis of multi-agent systems with intuitive Python interfaces. \textit{Verse} allows non-deterministic transitions and facilitates reachability analysis, offering a robust platform for analyzing the safety of complex autonomy algorithms. These tools not only ensure mathematical rigor but also provide actionable insights into system behavior under varying conditions~\cite{li2023refining, song2024verification}.

\textbf{Simulation-Based V\&V for Hybrid VTOL UAVs:} Simulation platforms play an essential role in the Verification and Validation (V\&V) of autonomy algorithms for hybrid VTOL UAVs, providing a controlled and repeatable environment to test performance of autonomy algorithms before real world deployment. Existing simulation-based V\&V solutions face significant limitations~\cite{chance2022determinism} when applied to hybrid VTOL UAVs. Few frameworks are able to provide high fidelity environments, so they face challenges to accurately verify vision-based perception algorithms, particularly under complex lighting, weather, or terrain conditions. Additionally, these platforms lack dedicated models for hybrid VTOL UAV dynamics, such as the transitions between hovering and forward flight. Few existing solutions provide comprehensive support for hybrid VTOL UAVs, highlighting a critical gap in current simulation-based V\&V. Addressing these shortcomings, this work leverages high fidelity photo-realistic simulator CARLA enhanced with VTOL-specific dynamics, MiniHawk (simulated using Gazebo)~\cite{minihawk_github} and NASA GUAM~\cite{guam_github}, to ensure robust V\&V of autonomy algorithms.

This paper aims to address these gaps by leveraging an advanced simulation environment integrated with simulation models of hybrid VTOL to enhance the robustness of simulation-based V\&V for autonomy algorithms. By focusing on the challenges specific to hybrid VTOL UAVs, this work contributes to bridging the sim-to-real gap and advancing the reliability of autonomy systems for future real-world applications.

\section{Simulation Environment}\label{sec:sim_env}

\begin{figure}[t]
  \centering
  \begin{minipage}[b]{0.48\textwidth}
    \centering    \includegraphics[width=.9\textwidth]{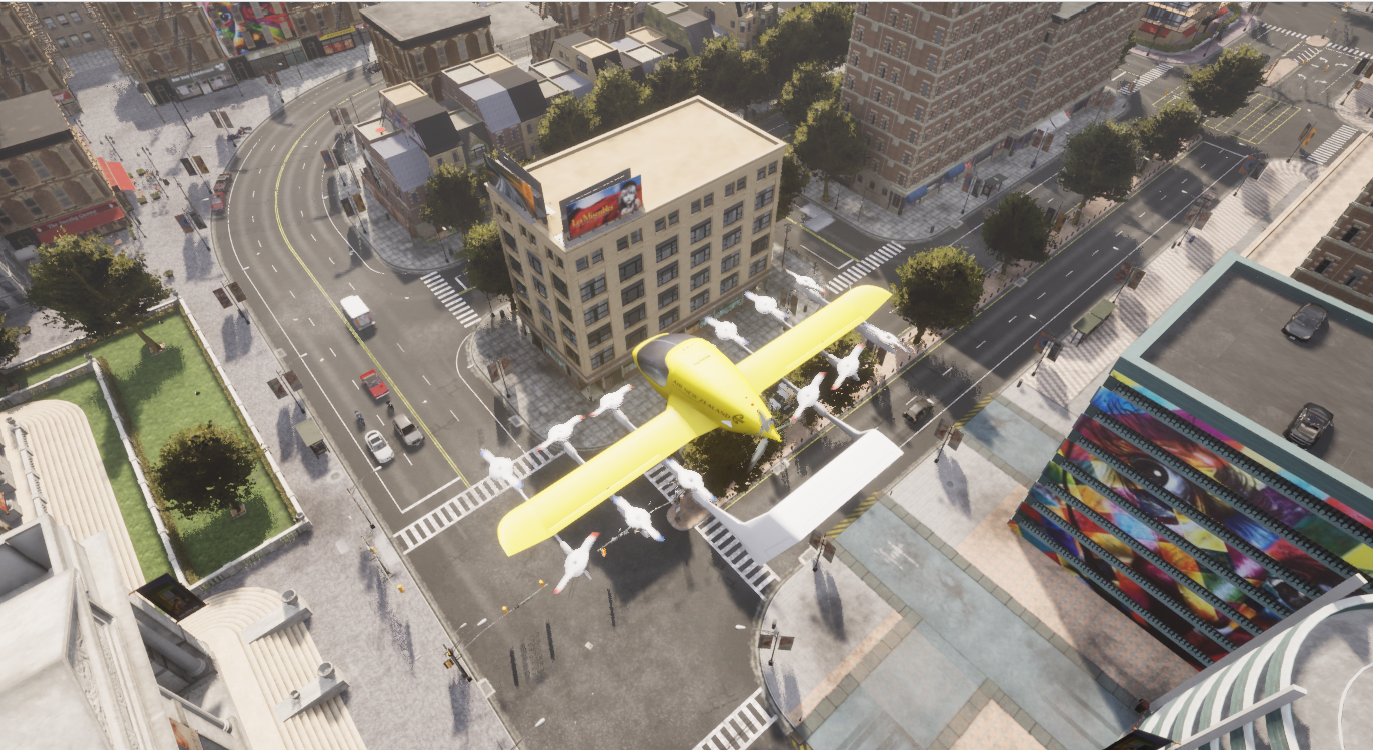}
    \caption{\label{fig:sim_screenshot}An overview of a simulation environment within CARLA with integrated air taxi.}
  \end{minipage}
  \hfill
  \begin{minipage}[b]{0.48\textwidth}
    \centering
    \includegraphics[width=\textwidth]{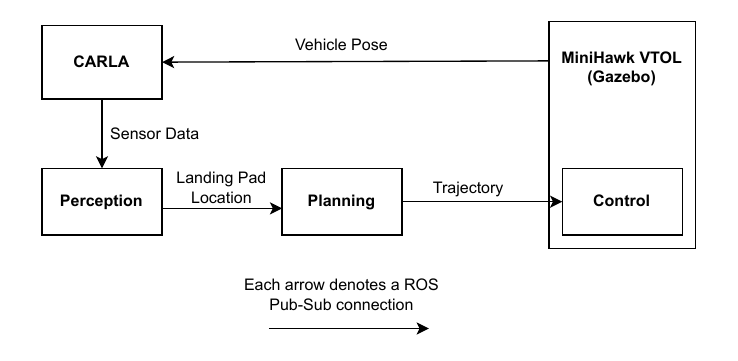}
    \caption{\label{fig:sim_blocks} An overview of the simulation loop for the automated landing system in air taxis.}
  \end{minipage}
\end{figure}

The simulation environment used in this work is designed to support high-fidelity simulations for autonomous air taxis operating in dynamic urban settings, as illustrated in Fig.~\ref{fig:sim_screenshot}. It builds on the popular urban environment simulator CARLA~\cite{dosovitskiy2017carla}, which has been extensively utilized for the verification and validation (V\&V) of autonomous ground vehicles in prior research~\cite{tahir2021intersection,won2022verification,zhang2023perception,chen2024end,goyal2024system}. CARLA, developed using Unreal Engine~\cite{EpicUE}, enables the creation of photorealistic 3D environments with realistic graphics and high-fidelity physics simulations. It also provides emulation of various sensors, including cameras and LiDAR, generating corresponding sensor data for the simulated environments. However, CARLA natively supports ground vehicles only.

To address this limitation, simulation environments that integrate aerial vehicle support in CARLA were developed~\cite{bansal2024synergistic}. This simulation environment integrates external high-fidelity physics engines to emulate the aerial vehicle. The simulation initially supported an advanced urban air mobility passenger vehicle~\cite{guam_github}, shown in Fig.~\ref{fig:sim_screenshot}.
To this, we add the MiniHawk-VTOL~\cite{minihawk_github,robowork_aerial_robotics}, a compact tiltrotor aerial vehicle with VTOL capabilities, that is the focus of this work.
MiniHawk's physics are simulated with high-fidelity in Gazebo~\cite{gazebo}, which computes the MiniHawk's pose.
The automated landing system evaluated in this work, as integrated within the simulation framework, is illustrated in Fig.~\ref{fig:sim_blocks}.
Details of the landing system design are provided in Section~\ref{ref:landing_system}. All components of the autonomous system communicate using the publisher-subscriber model facilitated by the Robot Operating System (ROS)~\cite{ros_noetic} middleware.

This framework leverages advanced simulation tools to create a robust and flexible testing environment that closely mimics real-world conditions, narrowing the sim-to-real gap from the simulation side. While Gazebo provides a graphical interface for visualizing the vehicle and its surroundings, the visualizations and corresponding sensor data lack realism, leading to a sim-to-real gap when using perception modules for decision-making. By relying on CARLA's high-fidelity simulation, the gap is significantly reduced as instead of using coarse images from Gazebo, we utilize photorealistic images generated by CARLA.

The simulation environment also offers extensive customization options for simulation scenarios. Configurable parameters include the initial pose of the ego vehicle\footnote{The ego vehicle refers to the vehicle that is autonomously controlled and observed in the simulation.}, the target landing pad location, weather conditions, wind conditions, vehicle parameters, and sensor configurations.
The scenarios and conditions used in this work are described in Section~\ref{sec:eval_scenario}.

\section{Verification and Validation Framework}

\subsection{Formal Verification with \textit{Verse}}

We are interested in determining whether the VTOL aircraft can land safely in time without colliding with any obstacles.
An execution of the VTOL landing system is the sequence of states that the aircraft travels starting from a certain initial state. For the VTOL landing system, given a set of initial states $X_0$, the reachable set at time t, denoted by $Reach(X_0, t)$, is the union of all possible executions starting from the initial set $X_0$ at time $t$.
Given the landing region $G$ and the unsafe set $U$ as the requirement for safe landing, the aircraft starting from $X_0$ can land safely at time $t_f$ without colliding with the obstacles if
\begin{align}
    & Reach(X_0, t_f) \subseteq G ~\text{and} \label{eq:landing_constraint}\\
    & Reach(X_0, t) \cap U= \emptyset, ~ \forall ~ 0\leq t\leq t_f \label{eq:collision_constraint}.
\end{align}
In this work, we employ the hybrid system verification tool Verse~\cite{li2023verse}, a Python-based library designed for verifying multi-agent hybrid scenarios, to analyze the VTOL aircraft system. Verse uses black-box simulators to describe system dynamics and implements the simulation-based reachability algorithm from DryVR~\cite{fan2017dryvr} to compute over-approximations of reachable sets with probabilistic accuracy guarantees. We integrate the simulation pipeline described in Section~\ref{sec:sim_env} as the black-box simulator for Verse.

\begin{figure}[ht]
    \centering
    \includegraphics[width=0.7\linewidth]{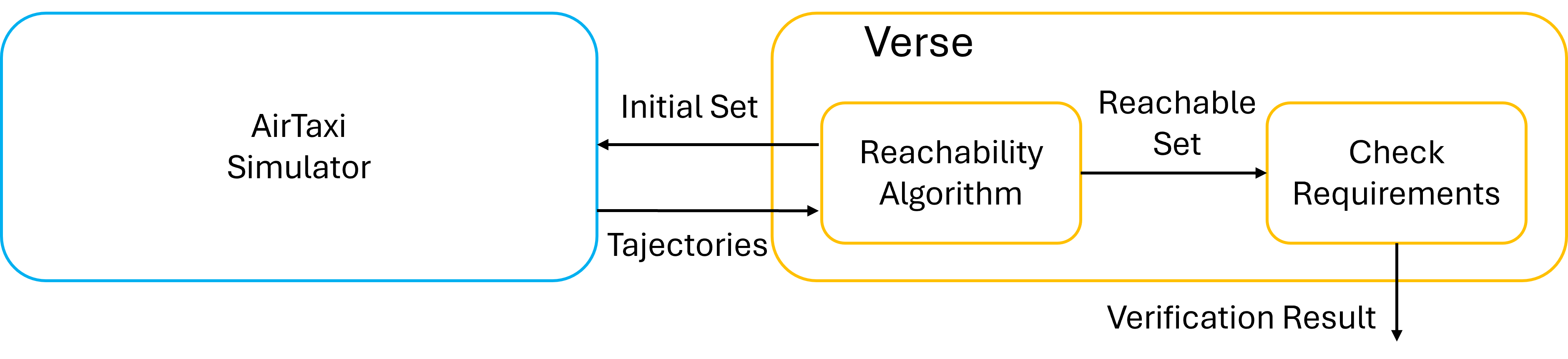}
    \caption{Verification Architecture for VTOL landing system.}
    \label{fig:verif_arch}
\end{figure}

Fig.~\ref{fig:verif_arch} illustrates the VTOL landing verification architecture. Starting from a prescribed set of initial conditions, Verse randomly samples initial states and generates corresponding trajectories using the simulation pipeline. These trajectories are then used to compute an over-approximation of the reachable set. By checking this reachable set against obstacle regions and desired goal sets, we can determine whether the system meets the specified safety and performance requirements.

\subsection{Validation Methodology and Bridging the Sim-to-Real Gap}

\begin{figure}[ht]
    \centering
    \begin{subfigure}[b]{0.6\textwidth}
    \includegraphics[width=\linewidth]{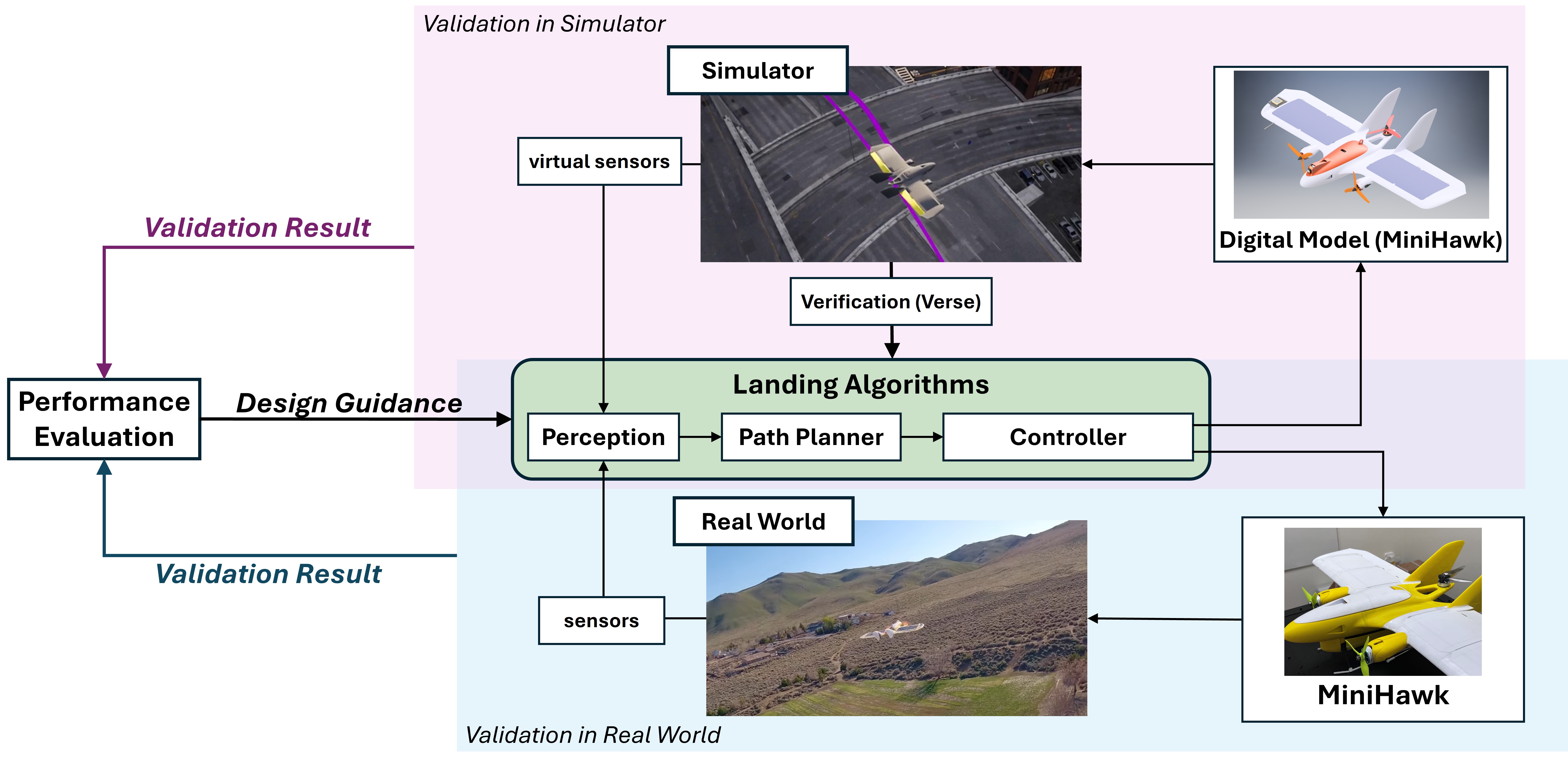}
    \caption{}
    \label{vnv-framework}
    \end{subfigure}
    \hfill
    \begin{subfigure}[b]{0.37\textwidth}
    \includegraphics[width=\linewidth]{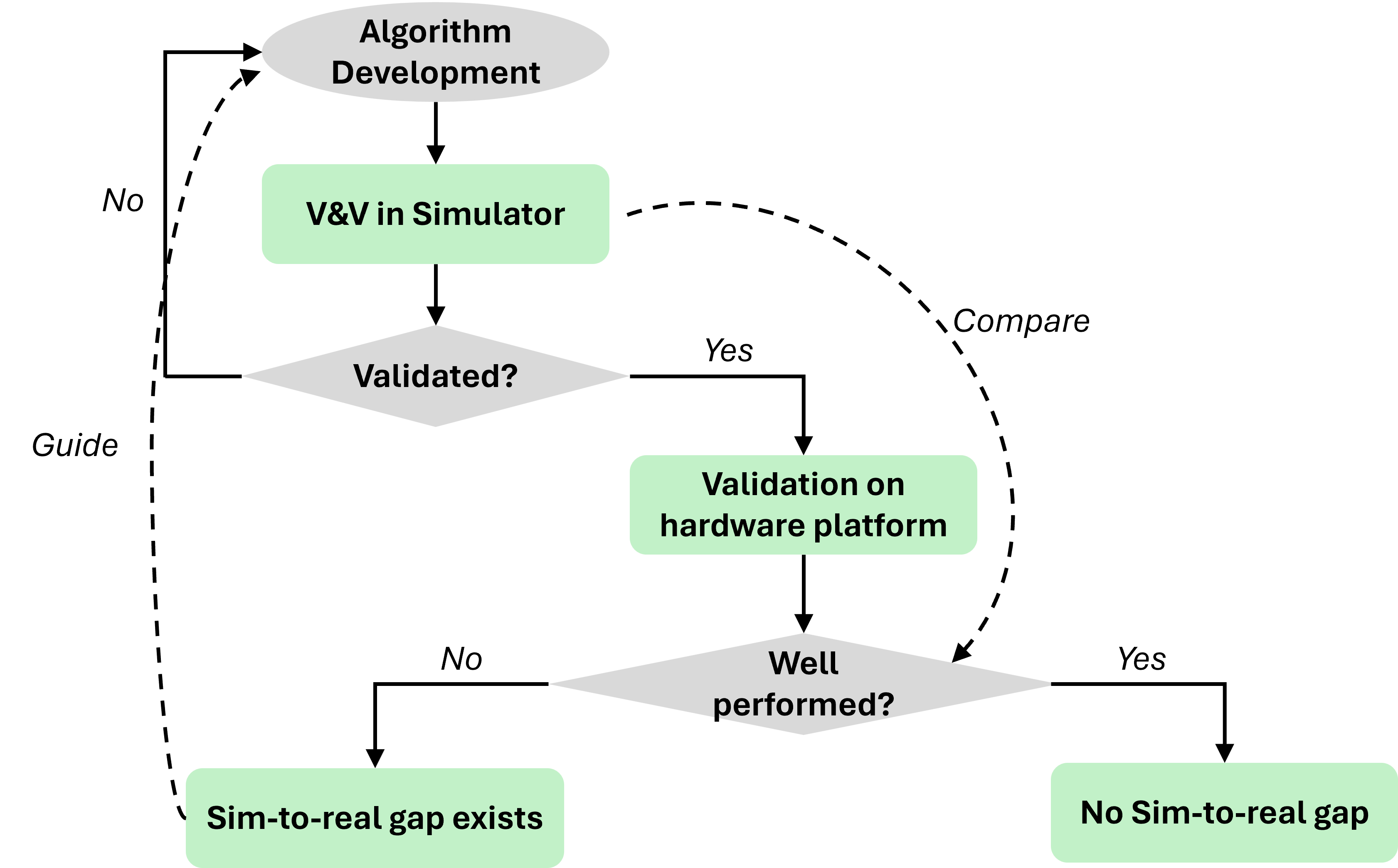}
    \caption{}
    \label{evaluation}
    \end{subfigure}
    \caption{A framework using simulated and real-world validation results to improve verified autonomy algorithms. }\label{vnv-framework-overall}
\end{figure}

Unlike verification, which establishes the correctness of algorithms under formal specifications, validation focuses on evaluating the performance of the system under diverse conditions to confirm their real-world applicability. For hybrid VTOL UAVs, this includes analyzing the ability to handle transition phase, navigate dynamic environments and adapt to disturbances. One challenge in validation is bridging the simulation-to-real (sim-to-real) gap. Simulators often simplify certain aspects of reality, such as sensor noises and environmental disturbances. These simplifications can lead to overestimation of algorithm performance and result in failures during hardware deployment. Given the complex dynamics and the requirement of outdoor experiments, this challenge is acute for hybrid VTOL vehicles. To address this problem, some real-to-simulation (real-to-sim) efforts are necessary, which involve incorporating characteristics from the real world into the simulator to make the simulation environment as close as possible to reality, thereby narrowing the sim-to-real gap. Furthermore, after obtaining validation results of the verified algorithms from simulator and real-world experiments, evaluating the performance differences can provide valuable feedback to guide the design and refinement of autonomy algorithms, contributing to further narrowing of the sim-to-real gap, as shown in Fig.~\ref{vnv-framework-overall}.

In this work, to narrow the sim-to-real gap, we adopt a multi-faceted approach:
\begin{itemize}
    \item Utilizing high-fidelity simulation environment developed on CARLA and enhanced with VTOL dynamics, to provide a repeatable environment for verifying the autonomy algorithms and testing.
    \item Using the model of the real-world vehicle platform (MiniHawk) in simulation.
    \item Training perception algorithms with real-world photos enhances robustness by exposing models to visual variations that simulators cannot fully imitate.
\end{itemize}

To further address the gap, we propose that the following approaches can help:
\begin{itemize}
    \item Integrating autonomy algorithms into real-time systems that combine hardware components with simulated environments to evaluate performance under realistic conditions (hardware-in-the-loop testing).
    \item Conducting controlled outdoor experiments that progressively increase in complexity, starting with simpler scenarios to identify and address potential failure points before deploying in fully operational environments.
\end{itemize}

These methods provide a framework for ensuring the correctness of autonomy algorithms for air taxis and enabling deployment in real-world applications.
\section{Case Study and Analysis}
\label{ref:case_study}

\subsection{Autonomous Landing System}

The simulation environment and the V\&V framework developed in this work are designed to be general and support a wide range of autonomy algorithms for VTOL UAVs. To demonstrate the effectiveness of the process and to facilitate a detailed analysis, we select the following specific perception, path planning, and control algorithms as representative case studies. These choices are motivated by their relevance to MiniHawk-VTOL and their alignment with the objectives of this study.
\label{ref:landing_system}
\subsubsection{Perception - Landing Zone (Helipad) Detection}~\label{casestudy:landingalgorithm:detection}

To ensure the safety and reliability of autonomous landing systems in hybrid VTOL UAVs, this work integrates a vision-based detection algorithm, using a YOLO (You Only Look Once)~\cite{redmon2017yolo9000} deep neural network (DNN) for helipad detection~\cite{SciTech2025_PerceptionTraining}. The system ensures precise identification and localization of the potential landing area and forms the basis for path planning.

The detection system is built on a dataset of annotated helipad images, which includes real-world images of helipads captured from aerial views from Google Earth images~\cite{bitoun2020helipadcat}. Helipad locations are marked in each image with bounding boxes, and the synthetic data augmentation includes varying lighting conditions (e.g., night) and
weather conditions (e.g., heavy rain), enhancing model robustness to real-world variations. The YOLOv8 model~\cite{ultralytics2023yolov8} is trained on this dataset to produce bounding boxes indicating helipad locations and the detection performance is evaluated using mean Average Precision at various Intersection-over-Union thresholds. Robustness and reproducibility are further ensured by disabling random data augmentations during controlled training cycles so that it allows for analysis of hyper-parameter effects. A Bayesian optimization framework is employed to ensure algorithm's robustness in dynamic conditions.
The helipad detection algorithm serves as an input to the path planning module. The center of the detected bounding box is utilized as the target end point for the planned trajectory to ensure accurate alignment of the UAV with the landing zone in the descending phase.

\begin{figure}[t]
    \centering
    \includegraphics[width=0.6\linewidth]{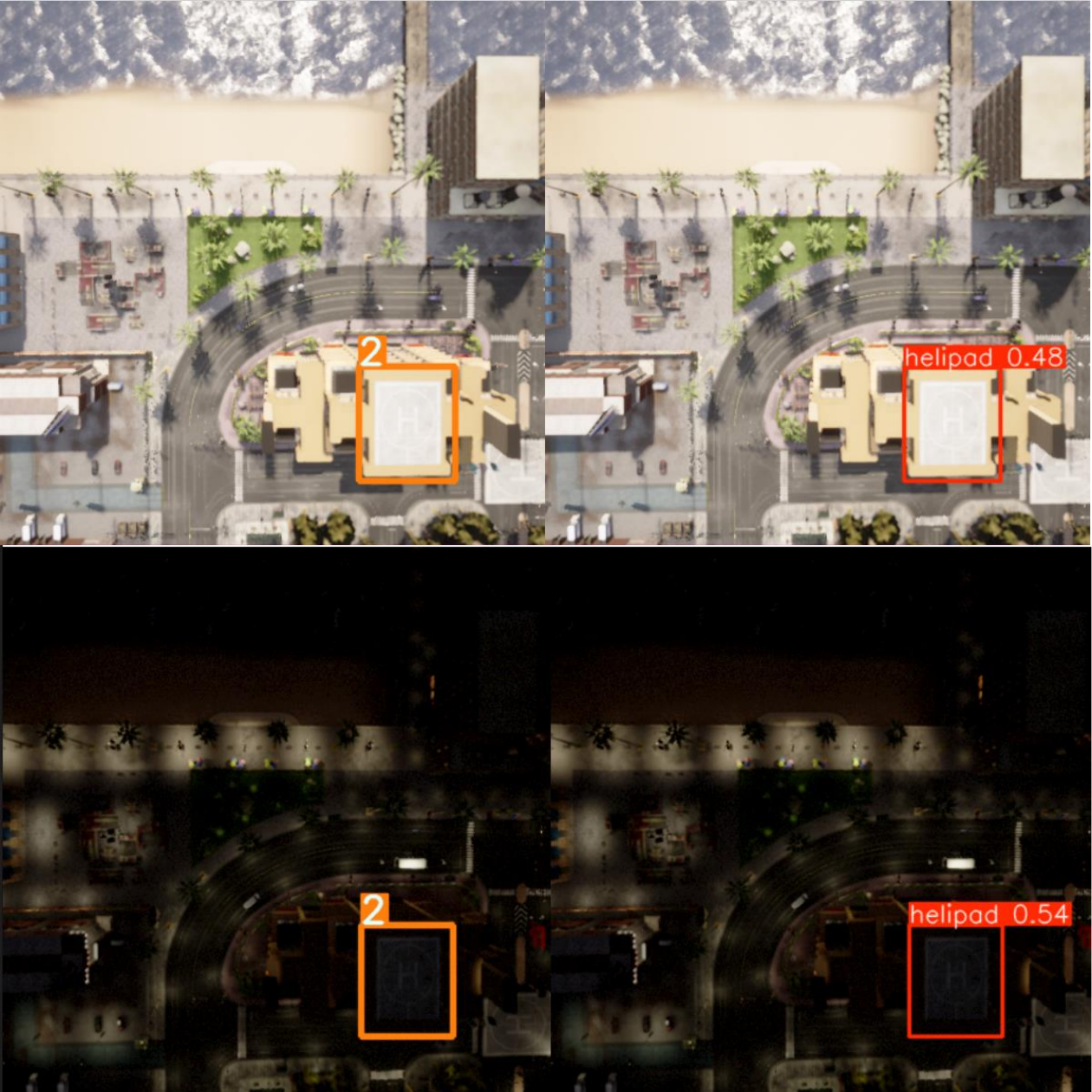}
    \caption{Examples of helipad detection under different lighting conditions, photos are taken in CARLA from a downward-facing RGB camera attached to the belly of the aircraft.}
    \label{fig:detect-diff-conditions}
\end{figure}

In this work, to accurately evaluate the effects of the perception on performance, we assume three perception setups: 1) a synthetic ideal perception system (Scenario1), 2) a synthetic perception system with bounded behavior (scenarios 2 and 3), and 3) YOLOv8 based perception system (scenarios 4 and 5).
These setups and scenarios are detailed in Section~\ref{sec:eval_scenario}.

Finally, while we evaluate the impact of obstacles between the aircraft and the landing pad, the detection of the obstacles itself is not a focus of this work.
In this work, the ideal obstacle information is assumed to be available.
Reliable obstacle detection and collision avoidance is an active area of research with significant prior solutions~\cite{masci2022method,bansal2022verifiable,shaikh2023self,bansal2024perception,bansal2024synergistic}.

\begin{figure}[t]
  \centering
  \begin{minipage}[t]{0.48\textwidth}
    \centering
    \includegraphics[width=.9\textwidth]{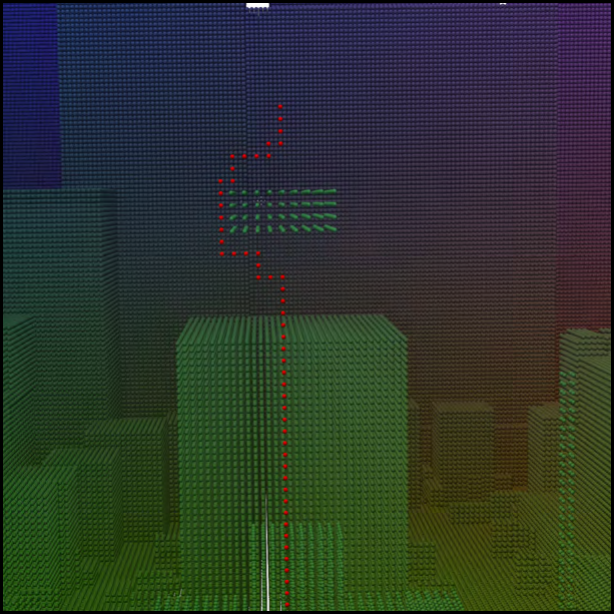}
    \caption{\label{fig:a_star_3d}Visualization of A* path planning in a 3D environment. The red dots indicate the planned trajectory. All other dots highlight obstacles.}
  \end{minipage}
  \hfill
  \begin{minipage}[t]{0.48\textwidth}
    \centering
    \includegraphics[width=0.9\textwidth]{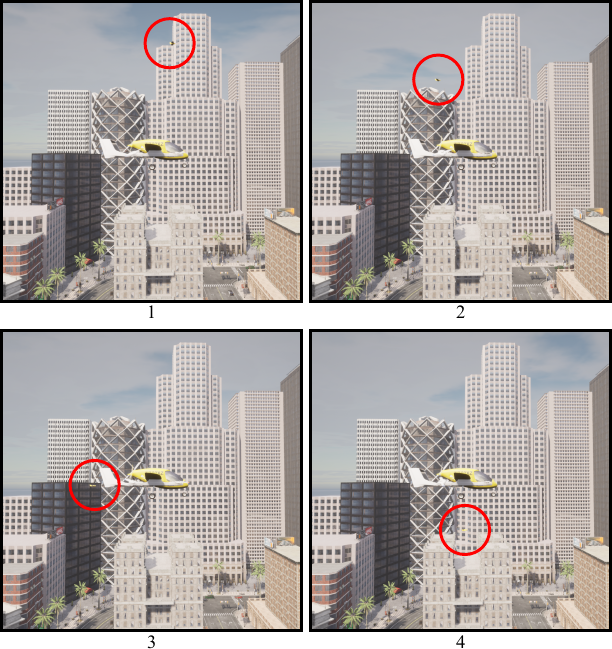}
    \caption{\label{fig:a_star_carla}Simulation of the A path planning algorithm to avoid another hovering aerial vehicle during landing. MiniHawk follows the trajectory depicted in Fig.~\ref{fig:a_star_3d}. The red circles indicate its position at different frames.}
  \end{minipage}
\end{figure}

\subsubsection{Path Planning - A* planner}
In this work, the A* algorithm is used to plan the path of VTOL aircraft navigating in a three-dimensional environment. A* is a graph-based search algorithm that determines the shortest path by minimizing the total cost function:

\begin{equation}
    f(n) = g(n) + h(n),
\end{equation}
where \(g(n)\) represents the actual cost from the start node to the current node \(n\), and \(h(n)\) denotes the heuristic estimate of the cost from \(n\) to the goal. The algorithm guarantees optimality under specific conditions. First, the heuristic \(h(n)\) must be admissible, meaning it does not overestimate the actual cost to the goal (\(h(n) \leq h^*(n)\), where \(h^*(n)\) is the true cost). Second, \(h(n)\) must be consistent, satisfying the inequality \(h(n) \leq c(n, n') + h(n')\) for any edge from \(n\) to \(n'\), where \(c(n, n')\) is the cost of the edge. These properties ensure the correctness and efficiency of the algorithm~\cite{hart1968formal}.

A* operates on a discretized 3D representation of the environment, where nodes correspond to spatial points, and edges define feasible transitions between them~\cite{lavalle2006planning}. The combination of the cost-to-go (\(g(n)\)) and the heuristic (\(h(n)\)) allows A* to effectively balance exploration and exploitation during the search process. However, the computational complexity of A* increases significantly with the dimensionality and density of the search space, which poses challenges for real-time implementations, especially on resource-constrained systems.

To mitigate these challenges, the action space of the VTOL is limited to six discrete motions: forward, backward, left, right, upward, and downward. By constraining the available actions, the branching factor of the graph is reduced, thereby decreasing the computational burden of the algorithm. This makes the implementation feasible for onboard systems with limited processing power. However, this simplification comes at the cost of path quality, as the resulting trajectories are often less smooth and may include abrupt transitions.

To address the difficulties in tracking such paths, robust control strategies are required to compensate for the lack of continuity in the planned trajectories. Despite these limitations, the proposed approach achieves a balance between computational efficiency and path-planning performance. This makes A* suitable for real-time applications in autonomous VTOL navigation, demonstrating its ability to provide efficient and reliable path planning under tight resource constraints.

Fig.~\ref{fig:a_star_3d}~ illustrates the operation of the A* algorithm in a 3D environment. The grid represents the discretized obstacles in the search space. The red points indicate the waypoints defining the trajectory generated by the path planner. All other points represent obstacles. Fig.~\ref{fig:a_star_carla} shows the resulting motion of the MiniHawk-VTOL around an existing obstacle --- another aerial vehicle.

\subsubsection{Controller}

For this case study, we utilize the ArduPilot flight stack as the primary controller for the MiniHawk in the simulation. The same ArduPilot has been implemented on the real MiniHawk using the mRo PixRacer Pro flight controller. This hardware-software combination ensures consistency in interface and behavior between simulation and physical platforms, significantly reducing the effort required to transplant algorithms verified in simulators onto the real vehicle. This streamlined transition facilitates an efficient and systematic V\&V process.
ArduPilot employs a PID-based control system that processes waypoint commands to manage motor inputs in a closed-loop manner. This conservative control approach prioritizes safety and robustness, particularly during outdoor experiments where environmental uncertainties can pose risks. Additionally, the tiltrotor functionality of the MiniHawk is natively supported by ArduPilot, enabling smooth transitions between hover and forward flight modes while maintaining control stability.

Future advancements may involve the integration of modern control strategies, such as the $\mathcal{L}_1$ adaptive control architecture~\cite{cao2008design}, to enhance the adaptability and performance under uncertain conditions. However, the primary challenges arise not from the algorithms themselves but from the risks inherent in hardware experiments. Low-level interfaces needed to control motor thrusts directly and bypass the safety constraints of waypoint-based system, demanding precise calibration and increasing the risk of instability during outdoor tests, where the margin for error is small.
By utilizing the PID-based controller in ArduPilot, safety and reliability are prioritized, providing a baseline for experiments. This choice enables us to focus on demonstrating the full V\&V framework while maintaining flexibility for integrating advanced control strategies in future work.

\subsection{Scenario Setting}
\label{sec:eval_scenario}

We consider emergency scenarios where MiniHawk must land on a helipad located on the top of a building.
The landing procedure in scenarios 1--3 begins immediately, with varying levels of prior knowledge about the target landing point from external sources (e.g., GPS), as described below. In scenarios 4 and 5, we assume no such information is available, relying solely on perception ($\S$\ref{casestudy:landingalgorithm:detection}) to detect the landing zone on the rooftop, with only the height of the landing pad known.
To evaluate the performance, correctness, and robustness of the autonomy algorithms under different conditions, we consider the following scenarios aimed at verifying the landing system's ability to handle uncertainties and land safely.
    In each scenario, the MiniHawk vehicle starts above the landing zone and navigates to the landing pad.
The center point of the landing pad on top of the building is $(0, 0, 29)$, with all dimensions in meters along the X, Y, and Z axes, respectively.

    \textbf{Scenario 1}: \emph{uncertainties in initial conditions with fixed landing point.}
    In this scenario, the MiniHawk initiates the landing procedure from varying initial positions.
    The landing point on the helipad is known and fixed at $(0, 0, 29)$.
    The initial position is within a bounded region above the landing pad $(\pm5, \pm5, 75 \pm 5)$.
    Within this bounded region the initial position is distributed uniformly along all axes.
    The uncertainty in initial position represents the variations in the trajectories with which the aerial vehicle may approach the landing pad.
    The fixed landing target mimics the conditions when an HD map of the landing location and error free GPS or GNSS\footnote{Global Positioning System (GPS), Global Navigation Satellite System (GNSS)} sensors are available to localize the aerial vehicle with respect to the landing target.

    \textbf{Scenario 2}: \emph{uncertainties in both initial conditions and landing points.}
    Building on the first scenario, this scenario introduces uncertainties in the location of the landing point. While the initial conditions remain the same as in Scenario 1, the target landing point is now within a region around the helipad's center, at a fixed height. The X and Y coordinates follow a normal distribution with $\mu = (0, 0, 29)$ and $\sigma = (1.5, 1.5, 0)$. The variance was chosen to ensure that the sampled target remains within the actual helipad.
    This mimics real-world conditions where the exact placement of the landing zone may vary.
    Specifically, this emulates the uncertainty in the landing pad position estimation by the sensors used to localize the vehicle, especially in the urban environments where such air taxi services are designed to operate in~\cite{cui2003autonomous,kos2010effects}.

    \textbf{Scenario 3}: \emph{shared airspace with intruder aircraft.}
    In this scenario, in addition to the conditions outlined in Scenario 2, a larger aircraft intrudes into MiniHawk's landing path, simulating a shared airspace environment with other aerial vehicles. The intruding aircraft is static at $(0, 0, 60)$, and has the following approximate dimensions: $\Delta x = 6.4\:\text{m},\:\Delta y = 10.8\:\text{m},\:\Delta z = 2.2\:\text{m}$. The initial position of MiniHawk is sampled from a uniform distribution within the range: $x \in [-5.5, -3.0]$, $y \in [-1.5, 1.5]$, $z \in [73.0, 77.0]$. The range of initial positions is intentionally reduced to enforce consistent behavior by MiniHawk when navigating around the obstacle, specifically ensuring it flies behind the intruding vehicle.
    This scenario highlights the integration of collision avoidance and safe landing capabilities in a complex, shared airspace environment.
    The reasons for the change in initial position generation, for this scenario and scenario 5, is discussed in Section~\ref{sec:discussion}.

    \textbf{Scenario 4}: \emph{uncertainties in initial conditions with vision-based landing pad detection.}
    This scenario is similar to Scenario 2, but the landing target is determined using vision-based landing pad detection, as described in Section~\ref{casestudy:landingalgorithm:detection}. The landing pad has a fixed height, which is assumed to be known. The X and Y coordinates of the landing target are defined as the midpoint of the bounding box detected by the perception system. Using the camera's intrinsic and extrinsic matrices, we convert this center point from camera coordinates to global coordinates and set it as the target. This scenario highlights the importance and impact of using real perception to detect the landing pad in the absence of ground-truth knowledge.

    \textbf{Scenario 5}: \emph{intruder aircraft with landing pad detection.}
    This scenario combines elements of Scenarios 3 and 4. Starting from scenario 3, the landing target is now determined by the perception system, while the same intruder aircraft from Scenario 3 obstructs the Minihawk ego vehicle's landing path. Notably, the intruder aircraft occludes the landing pad from the Minihawk's perspective.

\subsection{Reachability Analysis}

\begin{figure}[p]
    \centering
    \includegraphics[width=\linewidth]{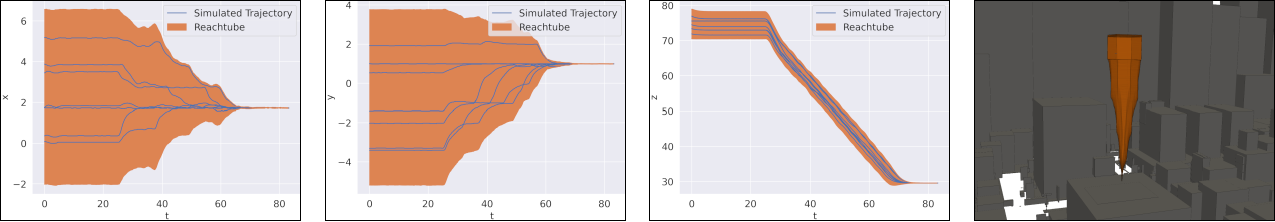}\caption{\label{fig:sc1_reach_1d}Reachability analysis for Scenario 1:
    uncertainties in initial conditions with fixed landing point.
    }
    \vskip 0.8cm
    \includegraphics[width=\linewidth]{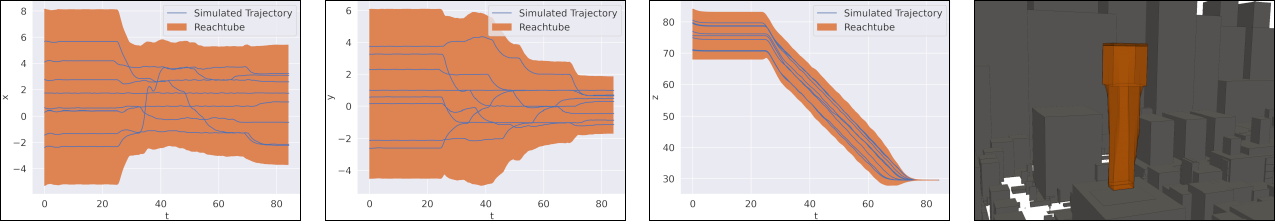}\caption{\label{fig:sc2_reach_1d}Reachability analysis for Scenario 2:
    uncertainties in both initial conditions and landing points.
    }
    \vskip 0.8cm
    \includegraphics[width=\linewidth]{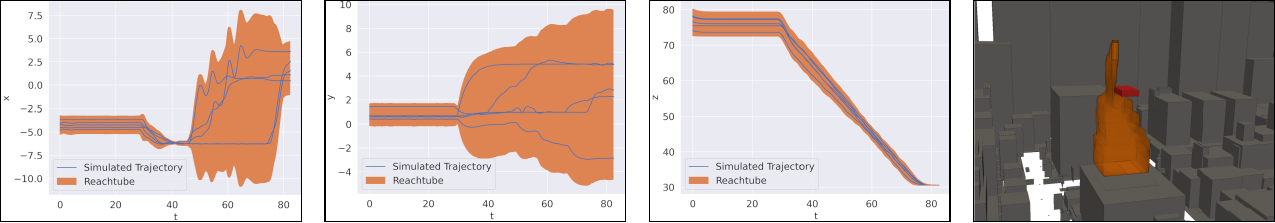}\caption{\label{fig:sc3_reach_1d}Reachability analysis for Scenario 3:
    shared airspace with intruder aircraft.
    }
    \vskip 0.8cm
    \includegraphics[width=\linewidth]{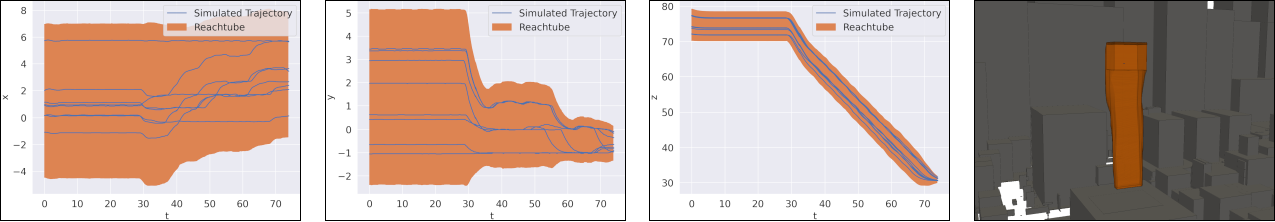}\caption{\label{fig:sc4_reach_1d}Reachability analysis for Scenario 4:
    uncertainties in initial conditions with vision-based landing pad detection.
    }
    \vskip 0.8cm
    \includegraphics[width=\linewidth]{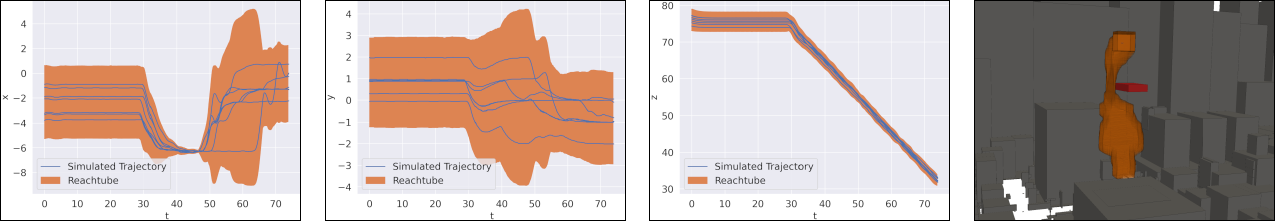}\caption{\label{fig:sc5_reach_1d}Reachability analysis for Scenario 5:
    intruder aircraft with landing pad detection.
    }
\end{figure}

We perform reachability analysis of all the scenarios described in Section~\ref{sec:eval_scenario} using hybrid system verification tool Verse and the results are discussed in this section. For all scenarios, Verse computes an over-approximation of the reachable set, which we term a \emph{reachtube}, up to a time horizon of $t_f=100$s. The reachable sets are computed from 10 trajectories simulated with a 0.25s time step.
Despite the high fidelity of the Minihawk simulation in Gazebo, inevitably, some differences exist. Rarely, the Minihawk simulation can destabilize, likely due to numerical error accumulation in the physics simulation. Such instances were manually identified and removed from the reachability analysis, as they do not reflect the behavior of the real vehicle in identical conditions.
With the computed reachtube, \emph{Verse can verify that the MiniHawk will land within the landing pad while avoiding obstacles}.
We now present and discuss the results for each scenario.

\textbf{Scenario 1}: \emph{uncertainties in initial conditions with fixed landing point.}
The simulated trajectories, reachable sets for each dimension, and a 3D visualization of the reachable set for Scenario 1 are shown in Fig.~\ref{fig:sc1_reach_1d}. It can be observed that all simulated trajectories are contained within the reachable set, providing evidence that the reachable set computed by Verse over-approximates the actual reachable set of the system. Despite initial uncertainty, the reachable set converges to a small region contained within the goal area. Additonally, the reachable set does not intersect with any of the gray boxes representing environmental obstacles, such as trees and buildings. These verification results confirm that MiniHawk can land safely without colliding with obstacles.

\textbf{Scenario 2}: \emph{uncertainties in both initial conditions and landing points.}
The results for Scenario 2 are shown in Fig.~\ref{fig:sc2_reach_1d}. From the plot, it is clear that uncertainties in the endpoint of the planned path, particularly those associated with the landing zone, cause the final portion of the reachable set in Scenario 2 to be larger than in Scenario 1. Despite this, the reachable set still converges toward the target region, demonstrating that the MiniHawk can successfully land on the targeted helipad on the building, even with uncertainty in the target location.

\textbf{Scenario 3}: \emph{shared airspace with intruder aircraft.}
This figure demonstrates that the reachable set of the MiniHawk does not intersect with the unsafe space occupied by the intruder (depicted in red in the 3D plot), indicating that the MiniHawk's path planner effectively avoids collisions with the aircraft intruding into the landing path. The reachable set for each dimension is also provided in Fig.\ref{fig:sc3_reach_1d}. From these plots, we observe that the collision avoidance behavior introduces greater uncertainties in the MiniHawk's motion, resulting in a reachable set with expanded volume, particularly along the y-axis.

\textbf{Scenario 4}: \emph{uncertainties in initial conditions with vision-based landing pad detection.}
The results for Scenario 4 are shown in Fig.~\ref{fig:sc4_reach_1d}. These results reveal that Verse computes an over-approximation of the actual reachable set, despite the uncertainties introduced by the perception pipeline. The outcome is similar to that of Scenario 2. However, in Scenario 4, unlike Scenario 2 where the radius of the reachable set in the z-dimension converges to zero, uncertainties persist across all x, y, and z dimensions at $t_f$ due to the uncertainty in the predicted landing point. The 3D plot further confirms that the reachable set does not intersect with any obstacles, demonstrating that the landing process is collision-free.

\textbf{Scenario 5}: \emph{intruder aircraft with landing pad detection.}
The reachability analysis for Scenario 5 is shown in Fig.~\ref{fig:sc5_reach_1d}. From these results, we observe that, despite the presence of an intruder and uncertainties from the perception algorithm, Verse successfully computes an over-approximation of the reachable set that covers the randomly simulated trajectories, with its volume remaining unchanged. Compared to the results from Scenario 3, the uncertainty in the z-dimension is larger at $t_f$. However, despite the additional uncertainty from perception, the radius of the reachable set in the y-dimension is smaller in Scenario 5 than in Scenario 3. This reflects that the error distribution of the centers of the detected bounding boxes is narrower than the normal distribution used in Scenario 3. In the final subfigure, we see that the reachable set avoids the intruder, demonstrating that MiniHawk can safely land without colliding with the intruding aircraft, even in the presence of perception uncertainties.

\subsection{Discussion}
\label{sec:discussion}

\emph{Reachability Analysis.}
The key result from the reachability analysis is that the goals of successful landing \eqref{eq:landing_constraint} and collision avoidance \eqref{eq:collision_constraint} are met.
Within the range of conditions and constraints described across the evaluation scenarios, the autonomous landing system is safe and reliable.

\emph{Dataset Size.}
The fineness and reliability of the reachable set, currently derived from 10 trials per scenario, could be improved by expanding the dataset to capture more trials and a broader range of conditions. While this would enhance confidence in the results, it would also demand increased computational resources. Given the limited performance of the onboard companion computer, optimizing the efficiency of the autonomy algorithms is crucial to support such expansions without compromising real-time capabilities.

\emph{Simulator Fidelity.}
Enhancing the simulation environment is equally important for improving validation. Efforts such as refining the digital dynamic model to better represent vehicle-specific characteristics and integrating more realistic environmental interactions can further narrow the sim-to-real gap. However, simulation-only V\&V faces inherent limitations. These challenges highlight the need for hardware-in-the-loop experiments and outdoor testing to identify failure modes that simulations may overlook.

\emph{Vision-based Refinement.}
The changes in reachability in scenarios 4 and 5 (Fig.~\ref{fig:sc4_reach_1d} and~\ref{fig:sc5_reach_1d}),
    as compared to scenarios 2 and 3 (Fig.~\ref{fig:sc2_reach_1d} and~\ref{fig:sc3_reach_1d}),
    motivate the inclusion of vision-based refinement to landing target position,
    especially in environments that cause GPS or GNSS sensor's accuracy to degrade.
While not a focus of this work, the vision-based refinement benefits from task aware learning and optimization~\cite{SciTech2025_PerceptionTraining}.

\emph{Over-approximation in V\&V Framework.}
A limitation of the V\&V framework is that the reachability analysis is necessarily a conservative overapproximation.
In this work, in Scenarios 3 and 5 ($\S$\ref{sec:eval_scenario}), had the Minihawk spawn logic been the same as other scenarios, the Minihawk would diverge all around the obstacle.
While these trajectories are safe, the resultant reachability analysis would erroneously include the region occupied by the obstacle.
This is a known limitation.
The common practice to addressing this challenge is partitioning the initial condition set, where the overall reachable set is obtained as the union of the reachable set from each partition. By refining the initial conditions, we can expect the resulting reachable set to avoid obstacles, thereby ensuring safety. However, in this paper, it is not achievable since the trajectories are pre-sampled.
Therefore, to present the reachability analysis for scenarios with obstacles, by modifying the spawn logic, the Minihawk is made to favor a subset of all possible trajectories around the obstacle.

\section{Conclusion and Future Work}

This work presents a Verification and Validation framework for vision-based landing systems in hybrid VTOL UAVs, focusing on the MiniHawk platform. By integrating Verse tool and high-fidelity photorealistic simulation environment built on CARLA, this work evaluates the safety and reliability of an autonomous landing system in cluttered urban environments.
Formal verification through reachability analysis provides mathematical guarantees for safe operation, while scenario-based validation assesses algorithm performance and robustness under various conditions. The findings highlight the effectiveness of combining high-fidelity simulation with formal verification tools in narrowing certain aspects of the sim-to-real gap and evaluating system performance. However, limitations emphasize the need for additional approaches such as hardware-in-the-loop testing and outdoor experiments, to refine and validate the algorithms.

The contribution of this study offers a foundation for advancing the reliability of autonomy algorithms in hybrid VTOL UAVs.
In future work, we would like to enhance the complexity of test scenarios by introducing dynamic intruders and complex environments. We want to enable the autonomy system to perform online planning using integrated perception and path planning algorithms. Additionally, we plan to address the heterogeneity of hybrid VTOL vehicle by designing new path planning and control algorithms that account for the distinct dynamics in different flight modes. We also aim to develop an interface for direct motor control, allowing us to implement and evaluate state-of-the-art controllers. Beyond the current focus on verifying the landing procedure, we intend to validate complete sequence including takeoff, forward flight and landing, to ensure performance across all mission phases. Furthermore, we plan to conduct real-world validation using the MiniHawk platform.

\section*{Acknowledgments}

This material is based upon work supported by
the National Aeronautics and Space Administration (NASA) under the cooperative agreement 80NSSC20M0229 and University Leadership Initiative grant no. 80NSSC22M0070, and
the National Science Foundation (NSF) under grant no. CNS 1932529 and ECCS 2311085.
Any opinions, findings, conclusions or recommendations expressed in this material
are those of the authors and do not necessarily reflect
the views of the sponsors.

We also thank Michael Acheson from NASA for sharing the GUAM simulator~\cite{guam_github}.

\bibliography{mybib}

\end{document}